\documentclass[english]{article}
\usepackage[T1]{fontenc}
\usepackage[latin1]{inputenc}
\usepackage{babel}
\usepackage{amsmath}
\usepackage{amsfonts}
\usepackage{mathenv}
\usepackage{graphicx}
\usepackage{subfigure}
\usepackage{authblk}
\numberwithin{equation}{section}
\newcommand{\email}[1]{\footnotesize\texttt{#1}}

\title{Double Relief with progressive weighting function}
\author{Gabriel Prat Masramon}
\author{Llu\'{i}s A. Belanche Mu\~{n}oz}
\affil{Faculty of Computer Science \\
	Polytechnical University of Catalonia \\
	Barcelona, Spain \\
	\email{\{gprat,belanche\}@lsi.upc.edu}
}
\date{15\textsuperscript{th} June 2006}

\begin{document}
\maketitle 
\begin{abstract}
Feature weighting algorithms try to solve a problem of great importance nowadays in machine learning: The search of a relevance measure for the features of a given domain. This relevance is primarily used for feature selection as feature weighting can be seen as a generalization of it, but it is also useful to better understand a problem's domain or to guide an inductor in its learning process. Relief family of algorithms are proven to be very effective in this task. 

On previous work, a new extension was proposed that aimed for improving the algorithm's performance and it was shown that in certain cases it improved the weights' estimation accuracy. However, it also seemed to be sensible to some characteristics of the data. An improvement of that previously presented extension is presented in this work that aims to make it more robust to problem specific characteristics. An experimental design is proposed to test its performance. Results of the tests prove that it indeed increase the robustness of the previously proposed extension. 
\end{abstract}

\section{Overview}
Feature selection is undoubtedly one of the most important problems in machine learning, pattern recognition and information retrieval, among others. A feature selection algorithm is a computational solution that is motivated by a certain definition of relevance.
However, the relevance of a feature may have several definitions depending on the objective that is looked after.

On the other hand, feature weighting algorithms try to estimate relevance (in the form  of weights to the features) rather than binarily deciding whether a feature is either relevant or not. This is a much harder problem, but also a more flexible framework from an inductive learning perspective. This kind of algorithms are confronted with the down-weighting of irrelevant features, the up-weighting of relevant ones and the problem of relevance assignment when redundancy is an issue. 

In this work we review Relief, one of the most popular feature weighting algorithms. Original Relief and some of its variants are presented on section \ref{seq:relief} drawing heavily on own earlier material.  Next, we revisit a "double" or feedback extension of the algorithm, that was firstly introduced in an own previous work, that takes its own estimations into account in order to improve general performance. Finally a new version of the algorithm is presented on section \ref{seq:drelief} that uses its own estimations in a progressive manner, it initially behaves like the traditional algorithm and gradually increases the importance of its estimates to behave at the end as the "double" version. An experimental design is presenten in secion \ref{seq:exp_design} to test the performance of the original algorithm versus the two proposed ones. Finally some results and conclusions are presented.

\section{Relief}\label{seq:relief}
Relief is a feature weighting algorithm that doesn't share one common characteristic of the feature selection and weighting methods. Most of them treat features individually assuming conditional independence of features upon the class. In the other hand, Relief takes all other features in care when evaluating a specific feature. Another interesting characteristic of Relief is that it is aware of contextual information being able to detect local correlations of feature values and their ability to discriminate from an instance of a different class. 

The main idea behind Relief is to assign large weights to features that contribute in separating near instances of different class and joining near instances belonging to the same class. The word "near" in the previous sentence is of crucial importance since we mentioned that one of the main differences between Relief and the other cited methods is the ability to take local context into account. Relief does not reward features that separate (join) instances of different (same) classes in general but features that do so for near instances. 

\begin{figure}[htbp]
Input: for each training instance a vector of feature values and the class value

Output: the vector W of estimations of the qualities of features
\begin{enumerate}
\item set all weights $W[A]:=0.0$; 
\item \textbf{for} $i:=1$ \textbf{to} $m$ \textbf{do begin} 
\item \hspace{2em}randomly select an instance $Ri$; 
\item \hspace{2em}find nearest hit $H$ and nearest miss $M$; 
\item \hspace{2em}\textbf{for} $A:=1$ \textbf{to} $a$ \textbf{do}  
\item \hspace{4em}$W[A]:=W[A] - \operatorname{diff}(A,Ri,H)/m + \operatorname{diff}(A,Ri,M)/m$
\item \textbf{end}; 
\end{enumerate}
\caption{Pseudo code of the original Relief algorithm}
\label{fig:relief}
\end{figure}

In Fig. \ref{fig:relief} we can see the original algorithm presented by Kira and Rendell in \cite{DBLP:conf/aaai/KiraR92}. We maintained the original notation that slightly differs from  the used above as now features (attributes) are labeled $A$. There we can see that in the aim of detecting whether the feature is useful to discriminate near instances it selects two nearest neighbors of the current instance $R_i$. One from the same class $H$ called the nearest hit and one from the different class $M$ (the original Relief algorithm only dealt with two class problems) called the nearest miss. With these two nearest neighbors it increases the weight ofthe feature if it has the same value for both $R_i$ and $H$ and decreases it otherwise. The opposite occurs with the nearest miss, Relief increases the weight of a feature if it has opposite values 
for $R_i$ and $M$ and decreases it otherwise.

One of the central parts of Relief is the difference function $\operatorname{diff}$ which is also used to compute the distance between instances as shown in Eq. \ref{eq:relief_distance}.
\begin{equation}
\delta(I_1,I_2)=\sum_{i}{\operatorname{diff}(A_i,I_1,I_2)}
\label{eq:relief_distance}
\end{equation}
The original definition of $\operatorname{diff}$ was an heterogeneous distance metric composed of the \textit{overlap} metric in Eq. \ref{eq:overlap} for nominal features and the 
normalized Euclidean distance in Eq. \ref{eq:euclidean} for linear features, which  \cite{DBLP:journals/jair/WilsonM97} called HEOM.
\begin{equation}
\operatorname{diff} (A,I_1 ,I_2 ) = 
\left.\{
	{\begin{array}{*{20}l}
	0 & \text{if }\operatorname{value}(A,I_1) = \operatorname{value}(A,I_2)  \\ 
	1 & \text{otherwise} 
\end{array}}  \right.
\label{eq:overlap}
\end{equation}
\begin{equation}
\operatorname{diff} (A,I_1 ,I_2 ) = \frac
	{\left| \operatorname{value} (A,I_1) - \operatorname{value}(A,I_2) \right|}
	{\max(A)-\min(A)}
\label{eq:euclidean}
\end{equation}
The difference normalization with $m$ guarantees that the weight range is [-1,1]. In fact the algorithm tries to approximate a probability difference in Eq. \ref{eq:dif_prob_relief}.
\begin{align}
\nonumber W[A] \approx & P(\text{different value of }A|\text{nearest instance from different class}) - \\
	& P(\text{different value of }A|\text{nearest instance from same class})
\label{eq:dif_prob_relief}
\end{align}
We can see that for a set of instances $\mathcal{I}$ having a set of features $\mathcal{F}$ this algorithm has cost $O(m \times |\mathcal{I}| \times |\mathcal{F}|)$ as it has to loop over $m$ instances. For each instance in the main loop it has to compute its distance from all other instances so we have $O(m \times |\mathcal{I}|)$ times the complexity of calculating $D_{Relief}$ and we can easily see from Eq. \ref{eq:relief_distance} that its complexity is  $O(|\mathcal{F}|)$, so we have our complexity:  $O(m \times |\mathcal{I}| \times |\mathcal{F}|)$. As $m$ is a user defined parameter we can in some measure control the cost of Relief algorithm having a tradeoff between accuracy of estimation (for large $m$) and low complexity of the algorithm (for small $m$). However $m$ can never be greater than $|\mathcal{I}|$.

\subsection{Extensions of Relief}
The first modification proposed to the algorithm is to make it deterministic by changing the outer loop through $m$ randomly chosen instances for a loop over all instances. This obviously increases the  algorithms computation cost which becomes $O(|\mathcal{I}|^2 \times |\mathcal{F}|)$ but makes  experiments with small datasets more reproducible. Kononenko uses this simplified version of the algorithm  in its paper \cite{DBLP:conf/ecml/Kononenko94} to test his new extensions to the original Relief. This version is also used by other authors \cite{DBLP:journals/ai/KohaviJ97} and its given the name \textit{Relieved}
with the final \textit{d} for "deterministic". 

We can find some extensions to the original Relief algorithm proposed in \cite{DBLP:conf/ecml/Kononenko94} in order to overcome some of its limitations: It couldn't deal with incomplete datasets, it was very sensible to noisy data and it could only deal with multi-class problems by splitting the problem into series of 2-class problems.

To able Relief to deal with incomplete datasets, i.e. that contained missing values, a modification of the $\operatorname{diff}$ function is needed. The new function must be capable of calculating the difference between a value of a feature and a missing value and between two missing values in addition to the calculation of difference between two known values. Kononenko proposed various modifications of this function in its paper and found one that performed better than the others it was the one in a version of Relief he called RELIEF-D (not to be confused with \textit{Releaved} mentioned above). The difference function used by RELIEF-D can be seen in Eq. \ref{eq:relief_d}.
\begin{equation}
\operatorname{diff}(A,I_1,I_2) =  
\begin{cases}
1 - P(value(A,I_2)|class(I_1)) & \text{ if $I_1$ is missing } \\
1 - \sum\limits_{a \in A}{[P(a|class(I_1)) \times P(a|class(I_2))]} & \text{ if both missing}
\end{cases}
\label{eq:relief_d}
\end{equation}

Now we will focus on giving Relief greater robustness against noise. This robustness can be achieved by increasing the number of nearest hits and misses to look at. This mitigates the effect of choosing a neighbor that would not have been the nearest without the effect of noise. The new algorithm has a new user defined parameter $k$ that controls the number of nearest neighbors to use. In choosing $k$ there is a tradeoff between locality and noise robustness. \cite{DBLP:conf/ecml/Kononenko94} states that 10 is a good choice for most purposes.

The last limitation was that the algorithm was only designed for 2-class problems. The straightforward extension to multi-class problems would be to take as the near miss the nearest neighbor belonging to a different class. This variant of Relief is the so-called Relief-E by Kononenko. But later on he proposes another variant which gave better results: This was to take the nearest neighbor (or the $k$ nearest) from each class and average their contribution so as to keep the contributions of hits and misses symmetric and  between the interval [0,1]. That gives the Relief-F (ReliefF from now on) algorithm seen in Fig. \ref{fig:relieff}.
\begin{figure}[htbp]
Input: for each training instance a vector of feature values and the class value

Output: the vector W of estimations of the qualities of features
\begin{enumerate}
\item set all weights $W[A]:=0.0$; 
\item \textbf{for} $i:=1$ \textbf{to} $m$ \textbf{do begin} 
\item \hspace{2em}randomly select an instance $R_i$; 
\item \hspace{2em}find $k$ nearest hits $H_j$;
\item \hspace{2em}\textbf{for} each class $C \neq class(R_i)$ \textbf{do}
\item \hspace{4em}find $k$ nearest misses $M_j(C)$; 
\item \hspace{2em}\textbf{for} $A:=1$ \textbf{to} $a$ \textbf{do}  
\item \hspace{4em}$W[A]:=W[A] - \sum\limits_{j=1}^k\operatorname{diff}(A,R_i,H_j)/(m\cdot k)+$
\item \hspace{6em}$\sum\limits_{C\neq class(R_i)} \left[\frac{P(C)}{1-P(class(R_i))} \\
\sum\limits_{j=1}^k\operatorname{diff}(A,R_i,M_j(C))\right]/(m\cdot k)$;
\item \textbf{end}; 
\end{enumerate}
\caption{Pseudo code of the ReliefF algorithm}
\label{fig:relieff}
\end{figure}

Relation to impurity functions, in specific with Gini-index gain can be seen in \cite{DBLP:journals/ml/Robnik-SikonjaK03} when developing the probability difference in Eq. \ref{eq:dif_prob_relief} in the case that the algorithm uses a large number of nearest neighbors (i.e., when the selected instance could be anyone from the set of instances). This version of the algorithm is called myopic ReliefF as it loses its context of locality property. Rewriting Eq. \ref{eq:dif_prob_relief} by removing the neighboring condition and by applying Bayes' rule, we obtain Eq. \ref{eq:dif_prob_relief_short}.
\begin{equation}
W'[A]=\frac{P_{samecl|eqval}P_{eqval}}{P_{samecl}} - \frac{(1-P_{samecl|eqval})P_{eqval}}{1-P_{samecl}}
\label{eq:dif_prob_relief_short}
\end{equation}
For sampling with replacement we obtain we have:
\begin{align*}
P_{eqval} & = \sum\limits_{c \in C}{P(c)^2} \\
P_{samecl|eqval} & = \sum\limits_{x \in X}{\left( \frac{P(x)^2}{\sum_{x \in X}{P(x)^2}} \times 
					\sum\limits_{c \in C}{P(c|x)^2}\right)}
\label{eq:dif_prob_relief_short}
\end{align*}

Now we can rewrite Eq. \ref{eq:dif_prob_relief_short} to obtain the myopic Relief weight estimation:
\begin{equation}
W'[A]=\frac{P_{eqval}\times GG'(X)}{P_{samecl}1-P_{samecl}}
\label{eq:myopic_relief}
\end{equation}
Where $GG'(A)$ is a modified Gini-index gain of attribute $A$ as seen in Eq. \ref{eq:modif_gini}.
\begin{equation}
GG'(X) = \sum_{x \in X}{\left(\frac{P(x)^2}{\sum_{x \in X}{P(x)^2}} \times \sum_{c \in C} {P(c|x)^2}\right)} - \sum_{c \in C} {P(c)^2 } 
\label{eq:modif_gini}
\end{equation}
As we can see the difference in this modified version from its original Gini-index gain is that Gini-index gain used a factor:
\begin{equation*} 
\frac{P(x)}{\sum_{x \in X}{P(x)}} = P(x)
\end{equation*} 
while myopic ReliefF uses:
\begin{equation*} 
\frac{P(x)^2}{\sum_{x \in X}{P(x)^2}}
\end{equation*}

So we can see how this myopic ReliefF in Eq. \ref{eq:myopic_relief} holds some kind of normalization for multi-valued attributes when using the factor $P_{eqval}$. This solves the bias of impurity functions towards attributes with multiple values. Anther improvement compared with Gini-index is that Gini-index
gain values decrease when the number of classes increase. The denominator of Eq. \ref{eq:myopic_relief} avoids this strange behavior.

\section{Double Relief}\label{seq:drelief}	
When more and more irrelevant features are added to a dataset the distance calculation of Relief degrades its performance as instances may be considered neighbors when in fact they are far from each other if we compute its distance only with the relevant features. In such cases the algorithm may lose its context of locality and in the end it may fail to recognize relevant features.

The $\operatorname{diff}(A_i , I_1, I_2)$ function calculates the difference between the values of the feature $A_i$ for two instances $I_1$ and $I_2$. Sum of differences over all features is used to determine the distance between two instances in the nearest hit and miss calculation (see Eq. \ref{eq:relief_distance}).

As seen in the k-nearest neighbors classification algorithm (kNN) many weighting schemes  which assign different weights to the features in the calculation of the distance  between instances (see Eq. \ref{weighted_distance}). 
\begin{equation}
\delta' (I_1 ,I_2 ) = \sum\limits_{i = 1}^a {w(A_i)\operatorname{diff}(A_i , I_1 ,I_2)}
\label{weighted_distance}
\end{equation}

In the same way that in \cite{DBLP:journals/air/WettschereckAM97} Relief's estimates of features' quality have been used successfully as weights for the distance calculation of kNN  we could use their estimation in the previous iteration to compute the distance between instances while searching the nearest hits and misses. We will refer to this version of ReliefF as double ReliefF or in short dReliefF. 

\subsection{Progressively weighted double Relief}
The problem using the weights estimates could be that in early iterations these estimations could be too biased to the first instances and could be far from the optimal weights. So, for small $t$, $W[A_i]$ is very different from $W[A_i]_t$.

What we want is to begin the distance calculation without using the weight estimates and then, as Relief's weight estimates become more accurate (because more instances have been taken into account), increase the importance of these weights in the distance calculation. Lets have a distance calculation like the one in Eq. \ref{progressively_weighted_distance}.
\begin{equation}
\delta (I_1 ,I_2) = \sum\limits_{i = 1}^a {f(W(A_i)_t,t)\operatorname{diff}(A_i , I_1 ,I_2)}
\label{progressively_weighted_distance}
\end{equation}

We would like a function $f:\mathbb{R} \times (0,\infty) \rightarrow \mathbb{R} $ such that:
  \begin{itemize}
  \item $f(w,t)$ is increasing with respect to $t$
  \item is continuous
  \item $f(w,0) = 1$
  \item $f(w, \infty) = w$
  \end{itemize}

\begin{figure}[p] 
   \begin {center}
   \includegraphics[scale=0.7]{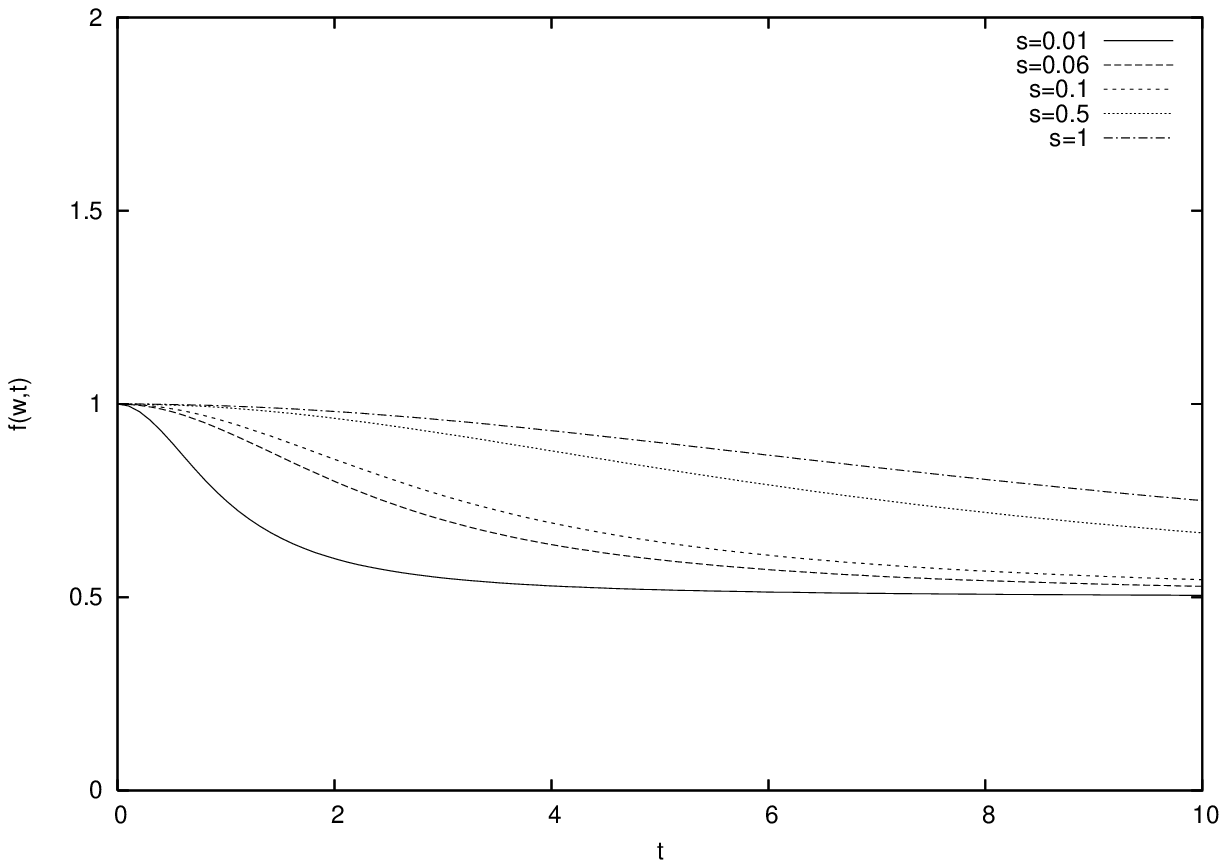}
   \end {center}
   \caption {Plot of function $f$ for 10 instances with $w=0.5$ and $c\left( t \right)=\left( t/m \right)^{2}$}
   \label{plotf_T}
\end{figure}

\begin{figure}[p] 
   \begin {center}
   \includegraphics[scale=0.7]{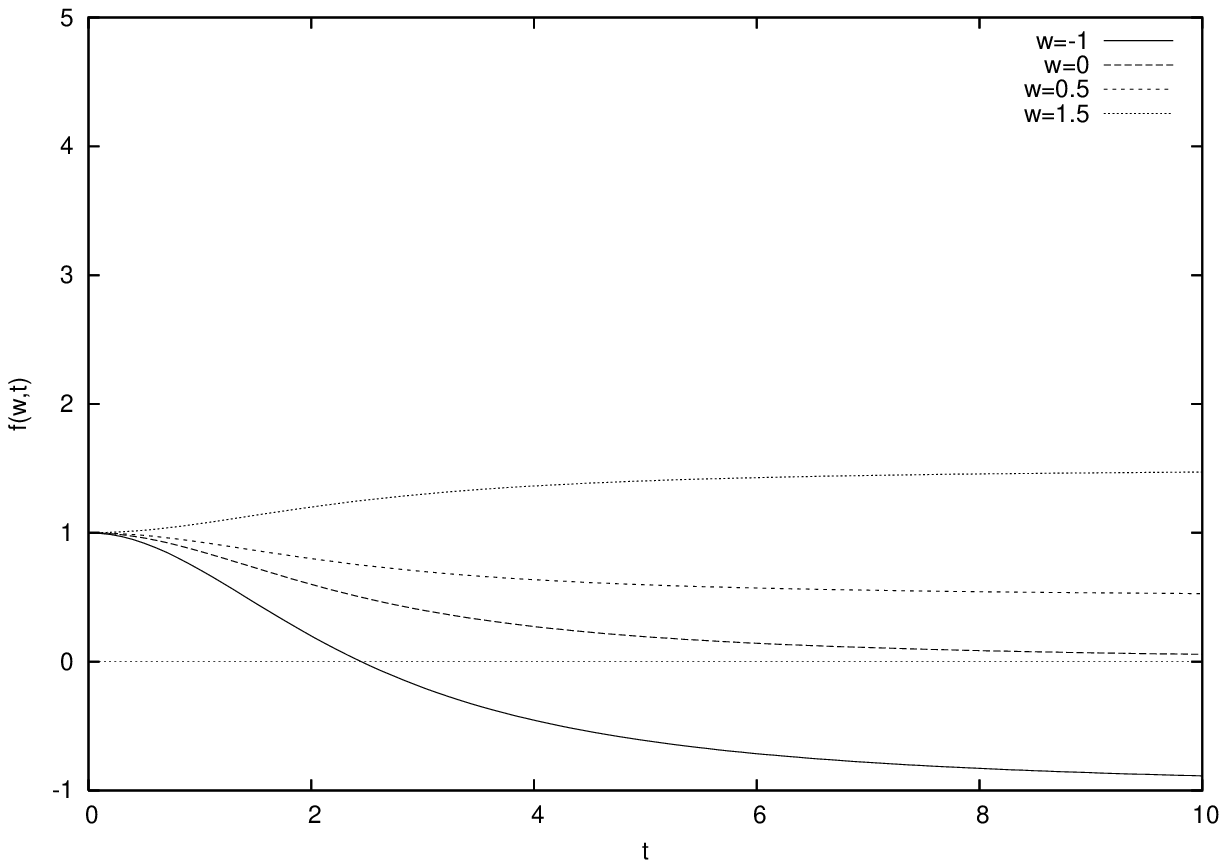}
   \end {center}
   \caption {Plot of function $f$ for 10 instances with $s=0.06$ and $c\left( t \right)=\left( t/m \right)^{2}$}
   \label{plotf_w}
\end{figure}

One such function could be the one in Eq. \ref{f}. And we will refer to the version of ReliefF using this distance equation as progressively weighted double relief or in short pdReliefF.
\begin{equation}
\begin{array}{l}
 f(w,t) = \frac{{(w - 1)c(t)}}{{c(t) + s}} + 1 \\ 
 \end{array}
\label{f}
\end{equation}
Where $s$ is a control parameter that determines the steepness and final value of the curve described by $f$ (see Fig. \ref{plotf_T}) and $c(t)$ is a function of the iteration number (e.g. $c(t)=t$). 
Another desirable property for our function would be that it always gives the same results regardless of the number of iterations. In other words, if $m$ is the total number of iterations, we would like $f(w,m)$ to be the same value whatever the value of $m$. To achieve that we must make $c(t)$ depend also on the total number of iterations $m$ so as to decrement the steepness of the function as the number of total iterations increases. A posible definition of $c(t)$ is shown in Eq. \ref{c}.
\begin{equation}
\begin{array}{l}
c\left( t \right)=\left( t/m \right)^{a}
 \end{array}
\label{c}
\end{equation}
In Fig. \ref{plotf_w} we can see how $f$ varies the influence of different weights (even a non realistic one that is greater than 1) as iterations go on. We can see that with high values of $s$ the function converges in the first few iterations and then it stabilizes its value near $w$ and for low values of $s$ it's value remains near 1 till the end. To choose a value we can compute the area left over and below the function. We can see the normal ReliefF as a particular case where $f(w,t)=1$ having maximum area and dReliefF as another particular case with $f(w,t)=w$ having minimum area. We want to choose the parameters to be in between the two. Specifically we could choose the parameters so as to leave 1/3 of the area below the function. For doing this we have to solve Eq. \ref{eq:area}
\begin{equation}
 \frac{\int_1^{m} f(w,t) \, dt-\int_1^{m} w \, dt}{\int_1^{m} 1 \, dt-\int_1^{m} w \, dt}=\frac{1}{3}
 \label{eq:area}
\end{equation}
A possible combination of parameters that solves the equation are: $a=2$ and $s=0.0633657\simeq0.06$. Graphicly it can be seen in Fig. \ref{plotf_w} that those values make weights' ponderations stay near 1 for half of the iterations and then takes values near the weights' values. This value has been chosen in our experiments.

\section{Experimental design}\label{seq:exp_design}
\subsection{Objective}
The above sections present three algorithms:
\begin{description}
 \item[ReliefF]
	The algorithm presented by Kononenko in \cite{DBLP:conf/ecml/Kononenko94}
 \item[dReliefF]
	The above algorithm using it's own partial weigts to ponderate attributes in distance calculation
 \item[pdReliefF]
	The above using a function to progressively increment the weights ponderation effect in distance calculation
 \end{description}
The objective of the experiments which will be presented is to compare performance of the three algorithms related to the factor of irrelevant attributes. The hypothesis is that the performance of the non-modified algorithm will be more affected by the number of irrelevant attributes increase due to their influence in distance calculation.

\subsection{Factors}
As stated before the key factor of the experiments is the ratio of irrelevant attributes, but there are some nuisance factors which have effect on the experiments' results. The factors considered in the experiments are:
\begin{itemize}
 \item Problem to solve
 \item Numeric vs. categoric attributes
 \item Number of relevant attributes
 \item Number of irrelevant attributes
 \item Data randomization
\end{itemize}
The main factor that will impact on performance results will be the problem we want to solve and in addition will be the most difficult to reduce. In order to eliminate it's influence, all the possible problems would have to be tried which is obviously impossible. Another factor that can clearly impact on performance is the type of the attributes as Relief has an heterogeneous function for distance calculation which depends on whether the attributes are numeric or categoric. So, to reduce the effect of these two factors the same experiments will be run on six different problems, three with numeric attributes and three with categoric ones. All the problems tested will be artificial to have sufficient knowledge about the data not to make performance of the weighting dependent on performance of a classifier.

Ranges for each factor have to be chosen. There has to be at least one relevant attribute and one irrelevant one in order to check whether the algorithm seems capable of distinguishing them, so both of them will start at 1 in our experiments. The number of irrelevant attributes will depend on the number of relevant ones in order to test with the same percentage of irrelevant attributes for each number of relevant attributes. A good choice could be to have at most twice the number of irrelevant attributes as the number relevant ones.

The upper bound for the number of relevant attributes will depend on the number of instances that are to be generated. It is interesting to test the algorithms with a wide range of attributes to instances ratios. We may arbitrarily set number of instances generated to 100. With that number of instances, it would be interesting to have at most 150 features for the ratio of attributes to instances not to get too low. If we want total features to keep below 150 with a number of irrelevant attributes of twice the number of relevant ones, we have to set upper bound to the number of relevant attributes to 50.

Finally 10 different sets of data will be generated for each combination of other factors to reduce the possible effect of randomly generating a pathologic set of data. 

\subsection{Design}
Here we have to decide which of all the possible combinations of factors will be tried in the experiments. The better way to reduce or eliminate the contribution to experimental error of each of the factors would be to treat them as blocking factors. That is to create homogeneous blocks in which the factors are kept constant while the target factor takes all its possible values. When blocking is not possible because of limited resources a random subset of each block can be run. 

With the ranges described above, there are a total of $3\times I\times N\times (N-1)$ different factor combinations for each problem as seen on Eq. \ref{eq:runs}, where $N$ is the number of relevant attributes and $I$ the number of iterations (i.e. random dataset generations) for each combination of relevant and irrelevant attribute numbers.
\begin{equation}
 \left(\sum\limits_{imp = 1}^N2imp\right)\times I_\text{iterations}\times 3_\text{algorithms}=3\times I\times N\times (N-1)
\label{eq:runs}
\end{equation}
That gives a total number of 76,500 different combinations for each problem.
With that number of combinations all combinations can be run. So the experimental design will be a full blocking design as shown on Fig. \ref{fig:exp_design} in an algorithmic way.
\begin{figure}[htbp]
\begin{enumerate}
\item \textbf{for each} $problem$ \textbf{in} $problems$ \textbf{do begin} 
\item \hspace{2em}\textbf{for} $impAtts:=1$ \textbf{to} $50$ \textbf{do begin} 
\item \hspace{4em}\textbf{for} $irrAtts:=1$ \textbf{to} $impAtts*2$ \textbf{do begin} 
\item \hspace{6em}\textbf{for} $iteration:=1$ \textbf{to} $10$ \textbf{do begin} 
\item \hspace{8em}execute problem with each algorithm; 
\item \textbf{end}; 
\end{enumerate}
\caption{Pseudo code of the experimental design}
\label{fig:exp_design}
\end{figure} 

\subsection{Problems}

\subsubsection{RDG1NamedContinuous}
A data generator that produces data randomly with numeric attributes by producing a decision list. The decision list consists of rules.  The rules have the form $c_x := \bigwedge\nolimits_1^n t$, where $t$ is an inequality term (i.e. $x<y$ or $x\geq{y}$) between some attribute and a random value. For each rule, the number $n$ will be a random number in the range $[1..10]$. An example set of rules can be seen on Eq. \ref{eq:rdg_cont}.
\begin{equation}
\begin{gathered}
\text{RULE 0: }c_0 := a_1 < 0.986 \wedge a_0 >= 0.65 \hfill \\
\text{RULE 1: }c_1 := a_1 < 0.95 \wedge a_2 < 0.129 \hfill \\
\text{RULE 2: }c_2 := a_1 >= 0.562 \hfill 
\end{gathered}
\label{eq:rdg_cont}
\end{equation}
Instances are generated randomly one by one. The class will be determined by the first rule that is true for the current instance. If decision list fails to classify the current instance, a new rule according to this current instance is generated and added to the decision list.
Irrelevant attributes are generated randomly in the range $[0,1]$.

\subsubsection{RandomRBFRandRed1}
\begin{figure}[htbp] 
   \begin {center}
   \includegraphics[scale=0.3]{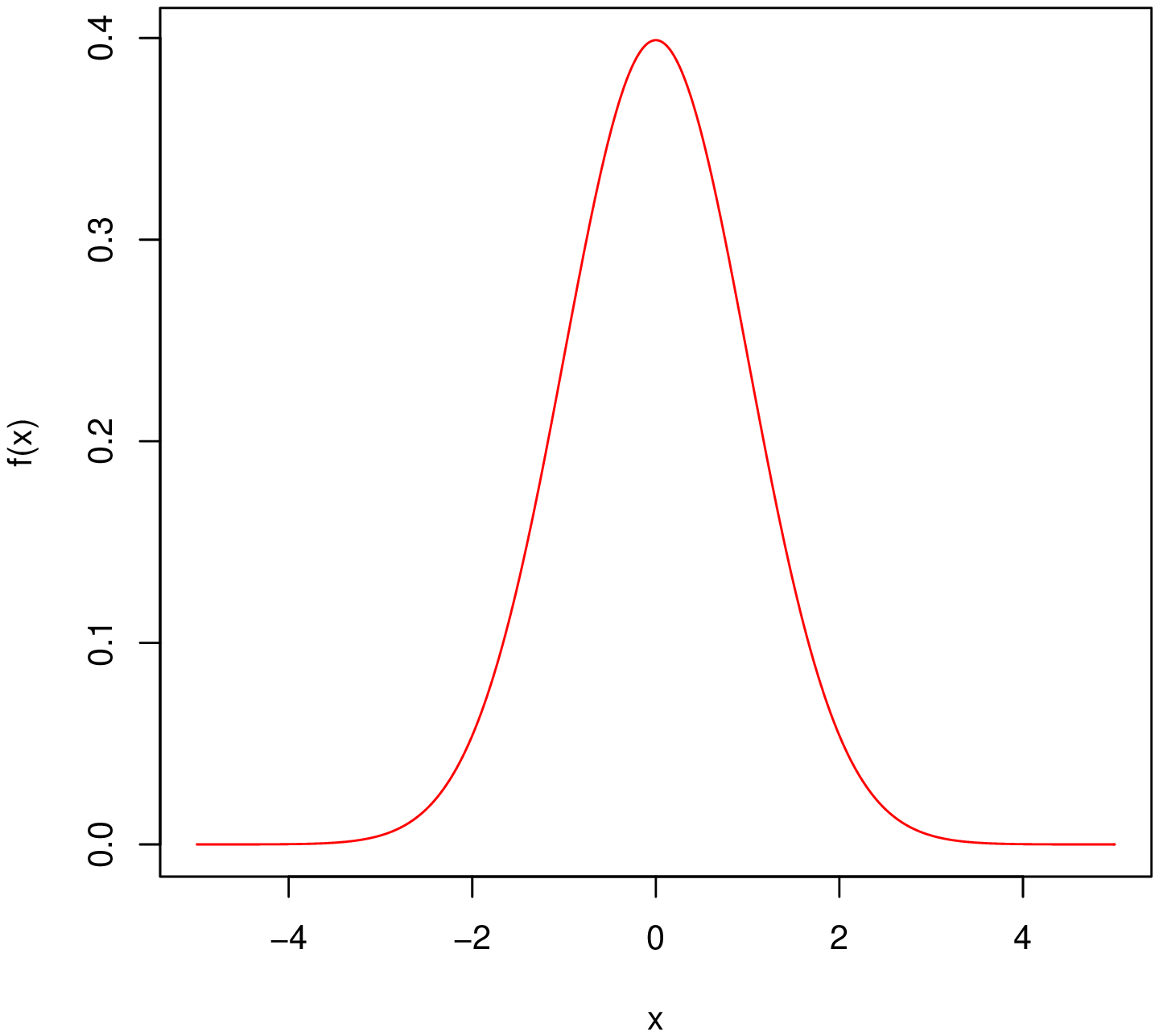}
   \end {center}
   \caption {Plot of function $f(x)$ with $\mu=0$ and $\sigma=1$}
   \label{fig:gauss}
\end{figure}

Radial basis functions (RBF) are functions which characteristic feature is that their response decreases (or increases) monotonically with distance from a central point. There are different formulas to describe the specific shape of the function and they usually have parameters to control the center and the distance scale. In this particular case, the function $f(x)$ used is the Gaussian which is described by Eq. \ref{eq:gauss} and can be seen on Fig. \ref{fig:gauss}. Its parameters are its mean $\mu$ and its standard deviation $\sigma$. A Gaussian RBF monotonically decreases with distance from the center.
\begin{equation}
f(x)=\frac{1}{\sigma\sqrt{2\pi}}\; \exp\left(-\frac{\left(x-\mu\right)^2}{2\sigma^2} \right)  
\label{eq:gauss}
\end{equation}

RandomRBF data is generated by first creating a random set of centers for each class. Each center is randomly assigned a weight, a central point per attribute, and a standard deviation. To generate new instances, a center is chosen at random taking the weights of each center into consideration. Attribute values are randomly generated and offset from the center, where the overall vector has been scaled so that its length equals a value sampled randomly from the Gaussian distribution of the center. The particular center chosen determines the class of the instance.
RandomRBF data contains only numeric attributes as it is non-trivial to include nominal values.
Irrelevant attributes are generated following the same Gaussian distribution for some random centers and standard deviation.
\subsubsection{NonMonotonic}
Let $r_a$ be a random value in the range $[0..1]$ to act as a ponderator for the attribute $a$. Now, for each instance $i$ generate a random value $r_i$ in the rage $[0..N]$, where $N$ is the number of important attributes. The value $a_i$ of the attribute $a$ for instance $i$ will be the one in Eq. \ref{eq:non_monotonic}.
\begin{equation}
a_i  = \left.\{
	{\begin{array}{*{20}l}
		{r_a \times r_i } & {{\text{if  }}i\mod 2 \ne 0}  \\
		{r_a \times \sqrt r _i } & {{\text{if  }}i\mod 2 = 0}  \\
	\end{array}} \right.
\label{eq:non_monotonic}
\end{equation}
The class for instance $i$ will be the integer part of $r_i$.
Irrelevant attributes are created randomly following a uniform distribution in the range $[0,1]$. 
\subsubsection{MajorityN}
Creates $n$ binary attributes and $i$ irrelevant attributes. The class attribute is 1 when the instance has a majority of 1s in the relevant attributes and 0 otherwise.
\subsubsection{ModuloP}
Each Modulo-p problem is described by a set $|\mathcal{R}|=n$ of relevant attributes and $i$ irrelevant attributes, both with integer values in the range $[0, p)$. The class $c$ can be defined as in Eq. \ref{eq:modulop}.
\begin{equation} 
c = \sum\limits_{r\in\mathcal{R}}{r\mod{p}}
\label{eq:modulop}
\end{equation}

\subsubsection{RDG1NamedCategoric}
The same data generator as for RDG1NamedContinuous but this time generating boolean attributes instead of numeric ones so now the rules are boolean predicates.

\section{Results}\label{seq:results}
\begin{figure}[p]
  \begin{center}
    \mbox{
	\subfigure[RDG1NamedContinuous]{
	\scalebox{0.3}{
	\includegraphics{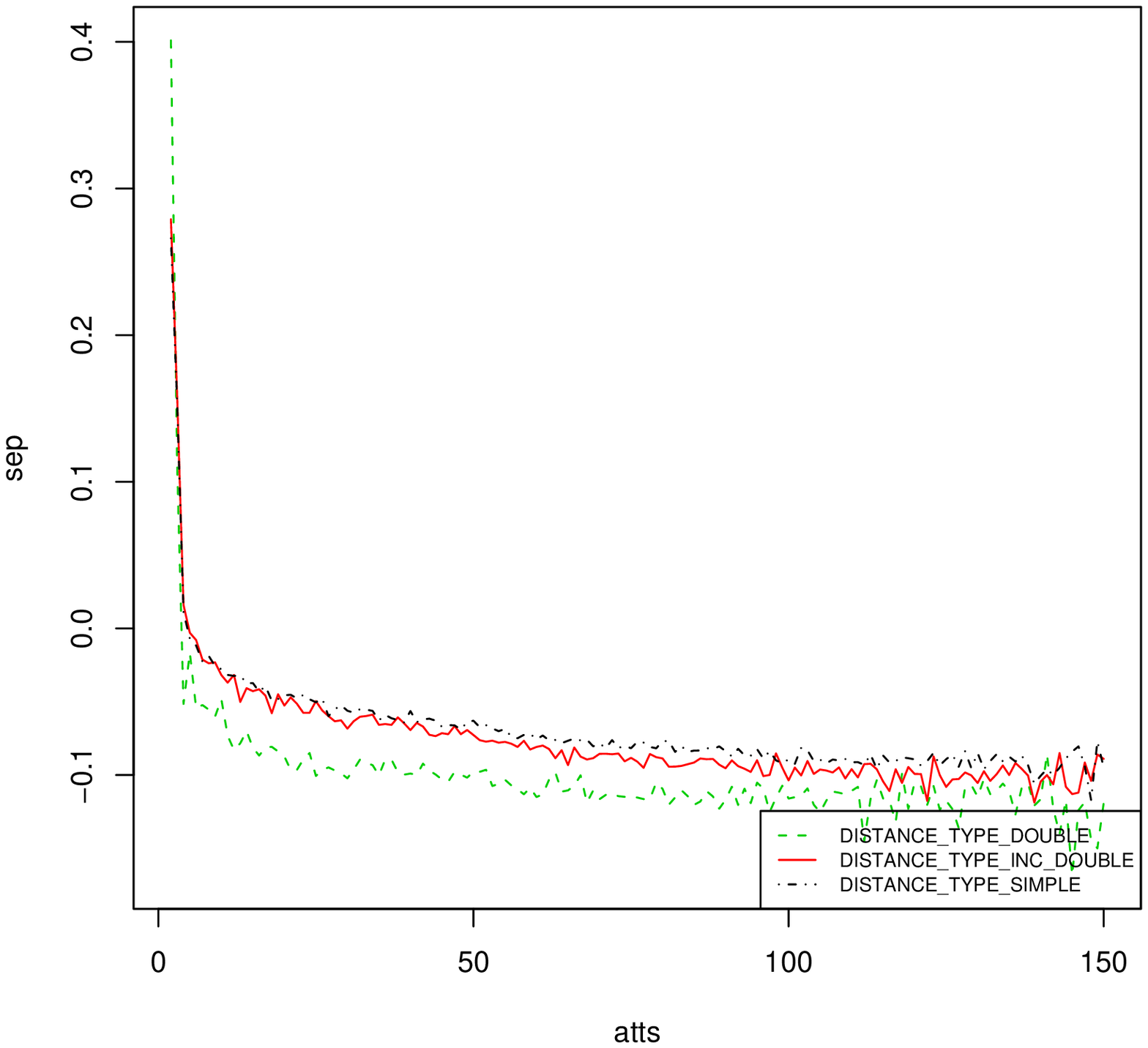}
	}}
	\subfigure[RandomRBFRandRed]{
	\scalebox{0.3}{
	\includegraphics{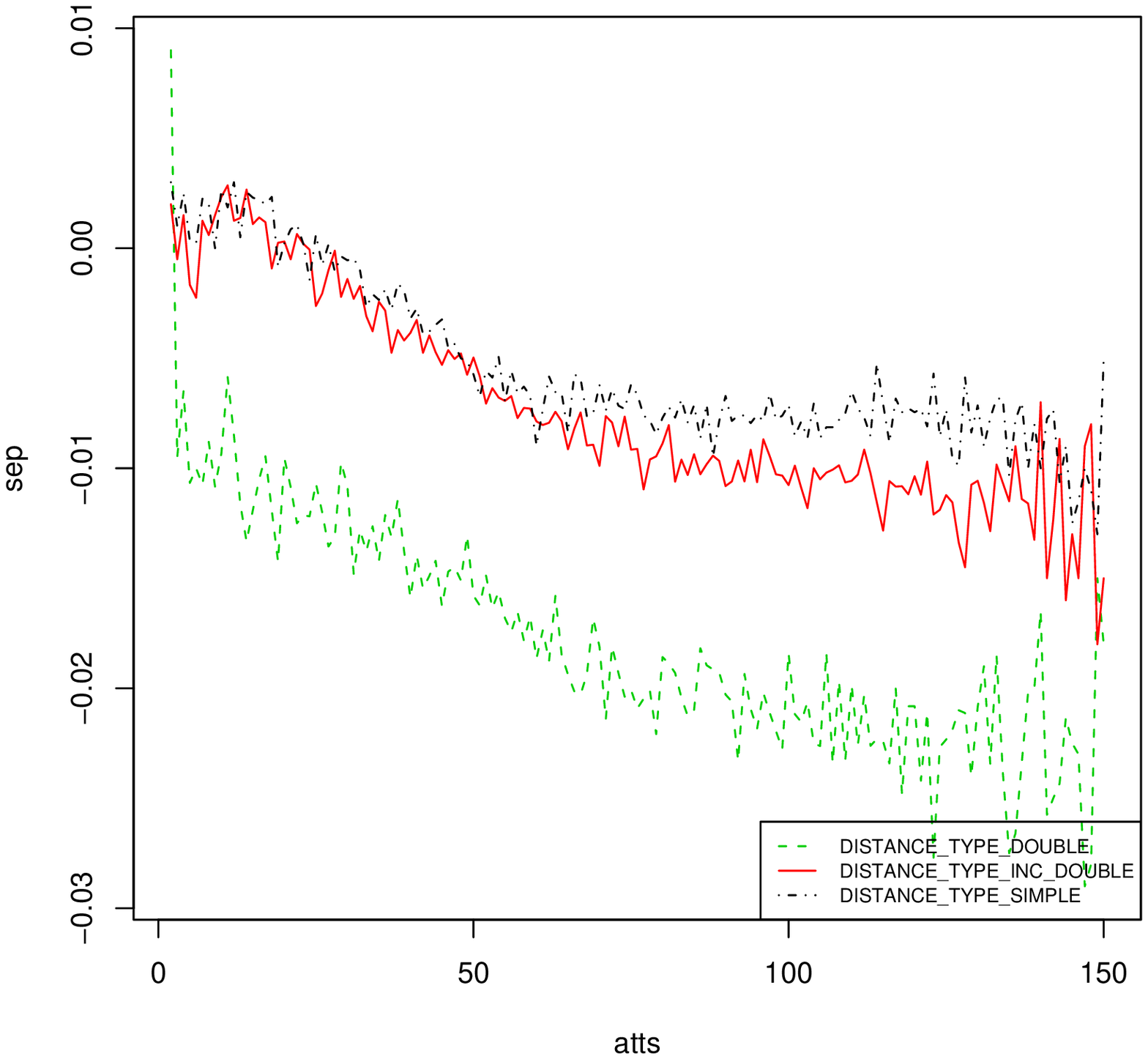}
	}}	
    }
    \mbox{
	\subfigure[NonMonotonic]{
	\scalebox{0.3}{
	\includegraphics{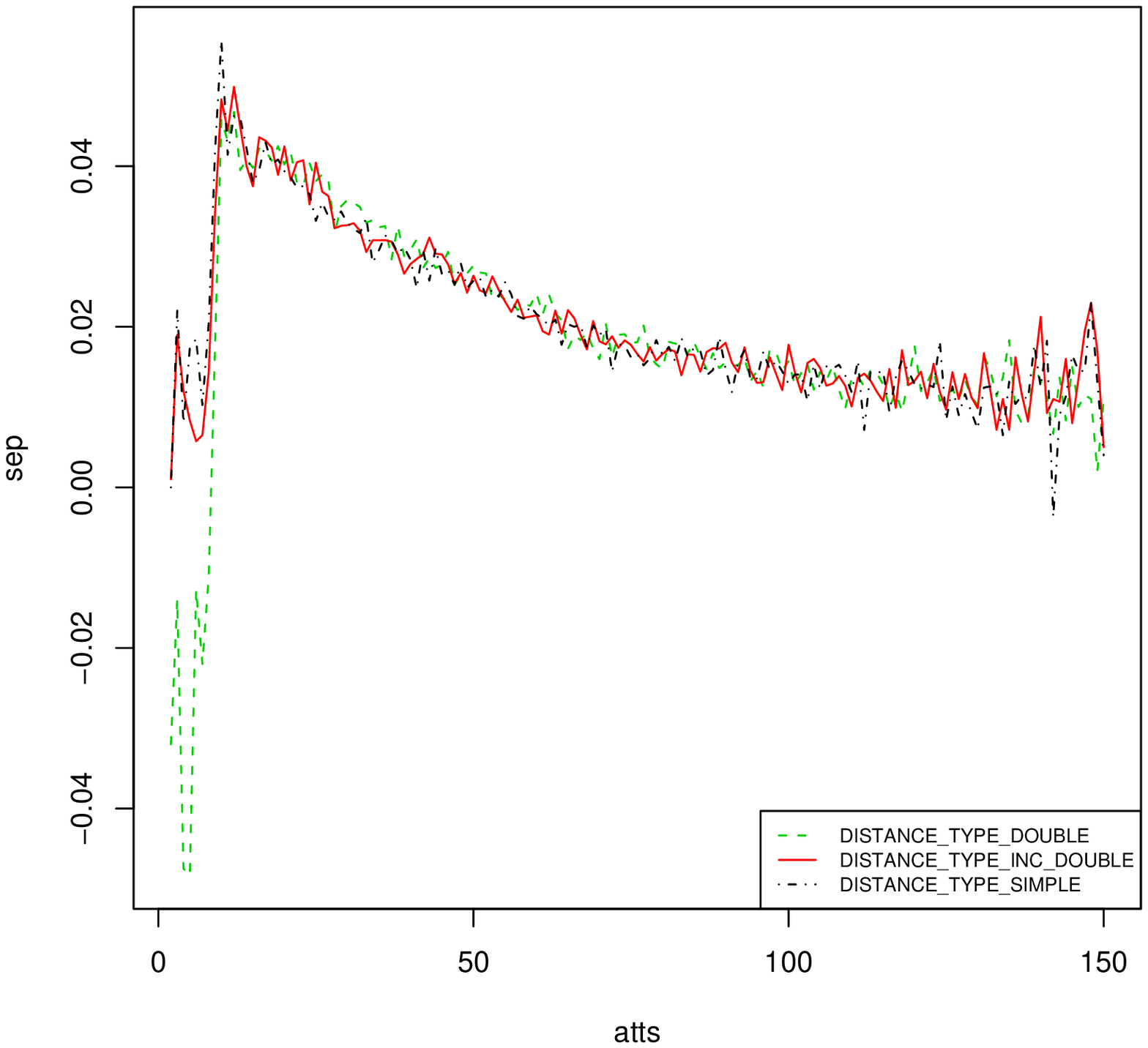}
	}}	
	\subfigure[MajorityN]{
	\scalebox{0.3}{
	\includegraphics{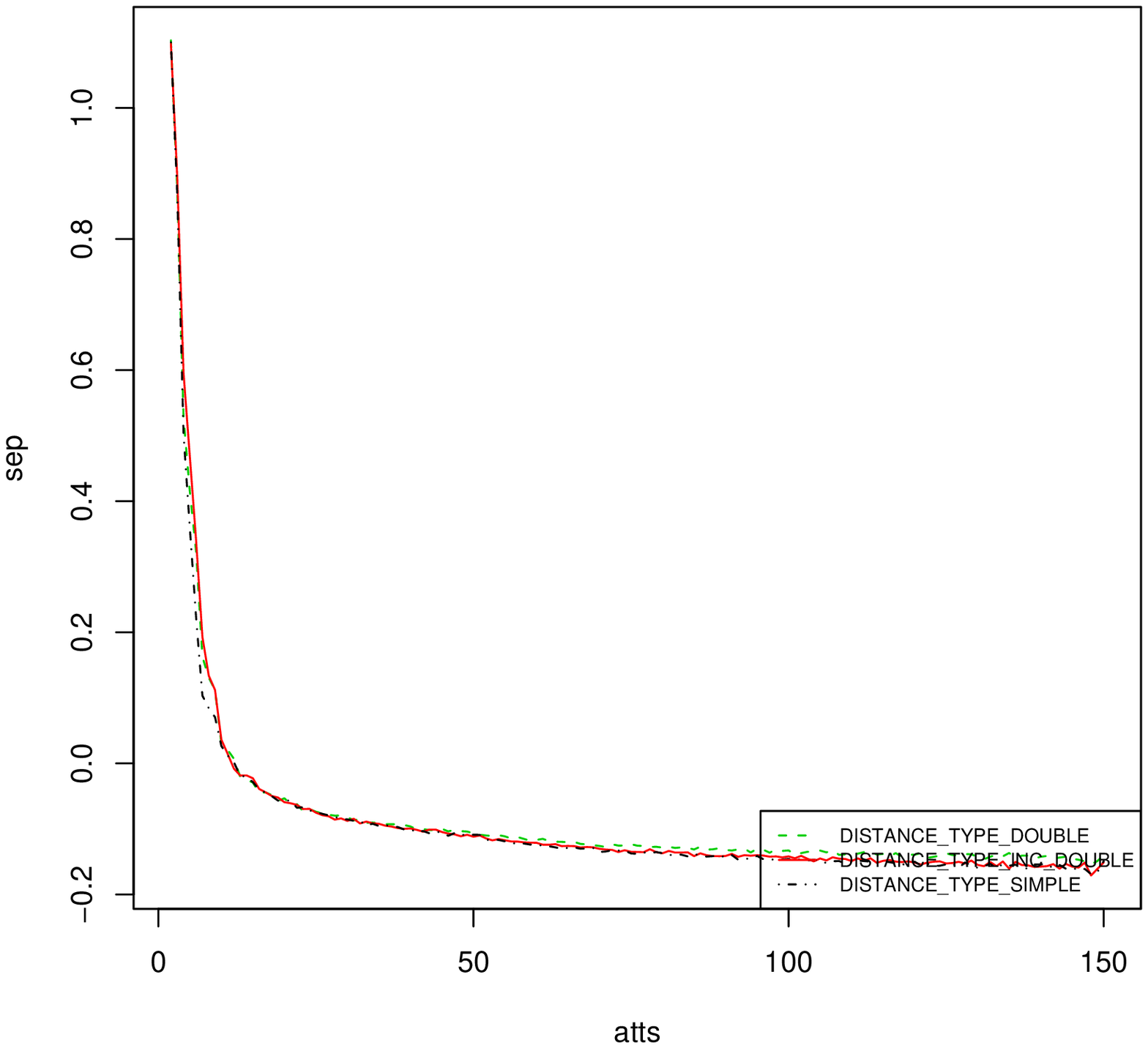}
	}}
    }
    \mbox{
	\subfigure[ModuloP]{
	\scalebox{0.3}{
	\includegraphics{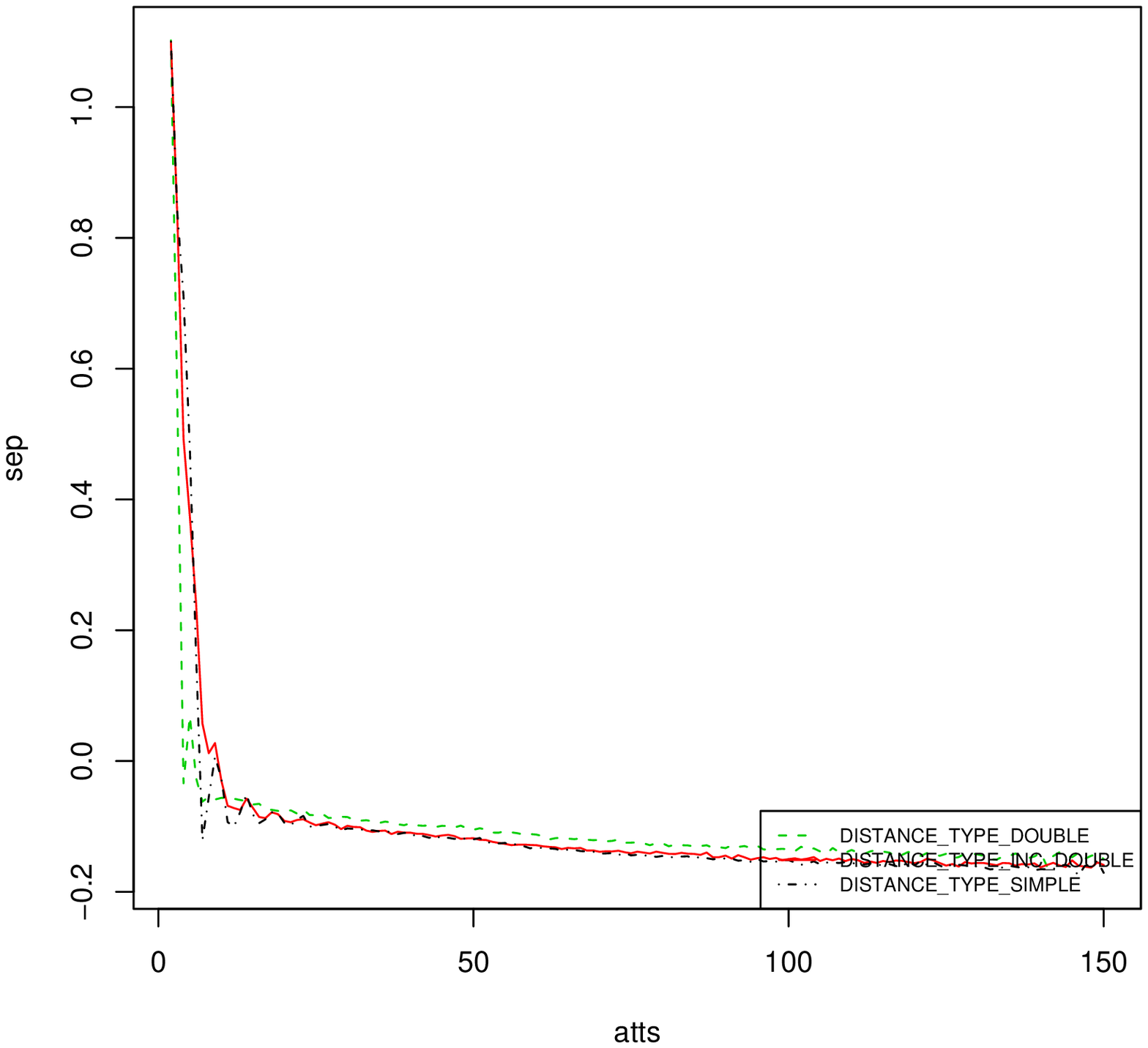}
	}}	
	\subfigure[RDG1NamedCategorical]{
	\scalebox{0.3}{
	\includegraphics{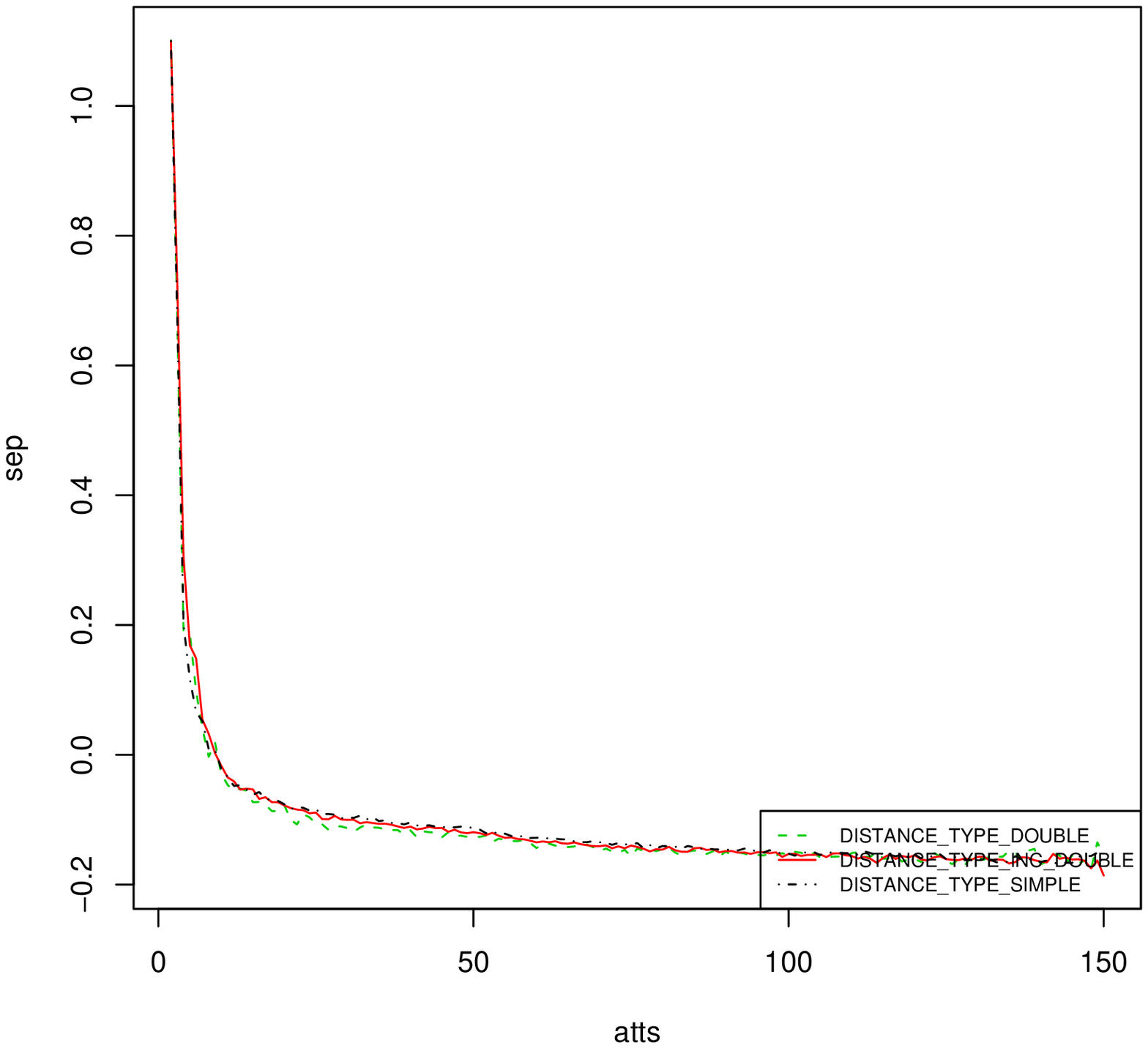}
	}}
    }
  \end{center}
  \caption {Separability versus total number of attributes for the three algorithms.}
  \label{fig:results}
\end{figure}

In this section the results of the above described experiments are presented. Six plots are presented in Fig. \ref{fig:results}. To clearly understand what the axes represent some notation has to be introduced.
Let $\mathcal{R}={r_1, r_2, \hdots , r_n}$ be the set relevant attributes and $\mathcal{I}={i_1, i_2, \hdots , i_m}$ the set of irrelevant ones having $\left|\mathcal{R}\right|=n$ and $\left|\mathcal{I}\right|=m$. And let $w(a)$ be the weight assigned by the algorithm to attribute $a$. Now, the x-axis represents the total number of attributes ($m+n$) and the y-axis the separability $s$ (i.e. the maximum weight assigned to a relevant attribute minus the maximum weight assigned to an irrelevant one). Formulas are shown in Eq. \ref{eq:results}.

\begin{equation}
\begin{gathered}
\text{x-axis: }m+n \hfill \\
\text{y-axis: }s = \left( \max\limits_{a_r\in\mathcal{R}}{w(a_r)} \right) - \left( \max\limits_{a_i\in\mathcal{I}}{w(a_i)} \right) \hfill
\end{gathered}
\label{eq:results}
\end{equation}

\begin{figure}[p] 
  \begin{center}
    \mbox{
	\subfigure[RDG1NamedContinuous]{
	\scalebox{0.3}{
	\includegraphics{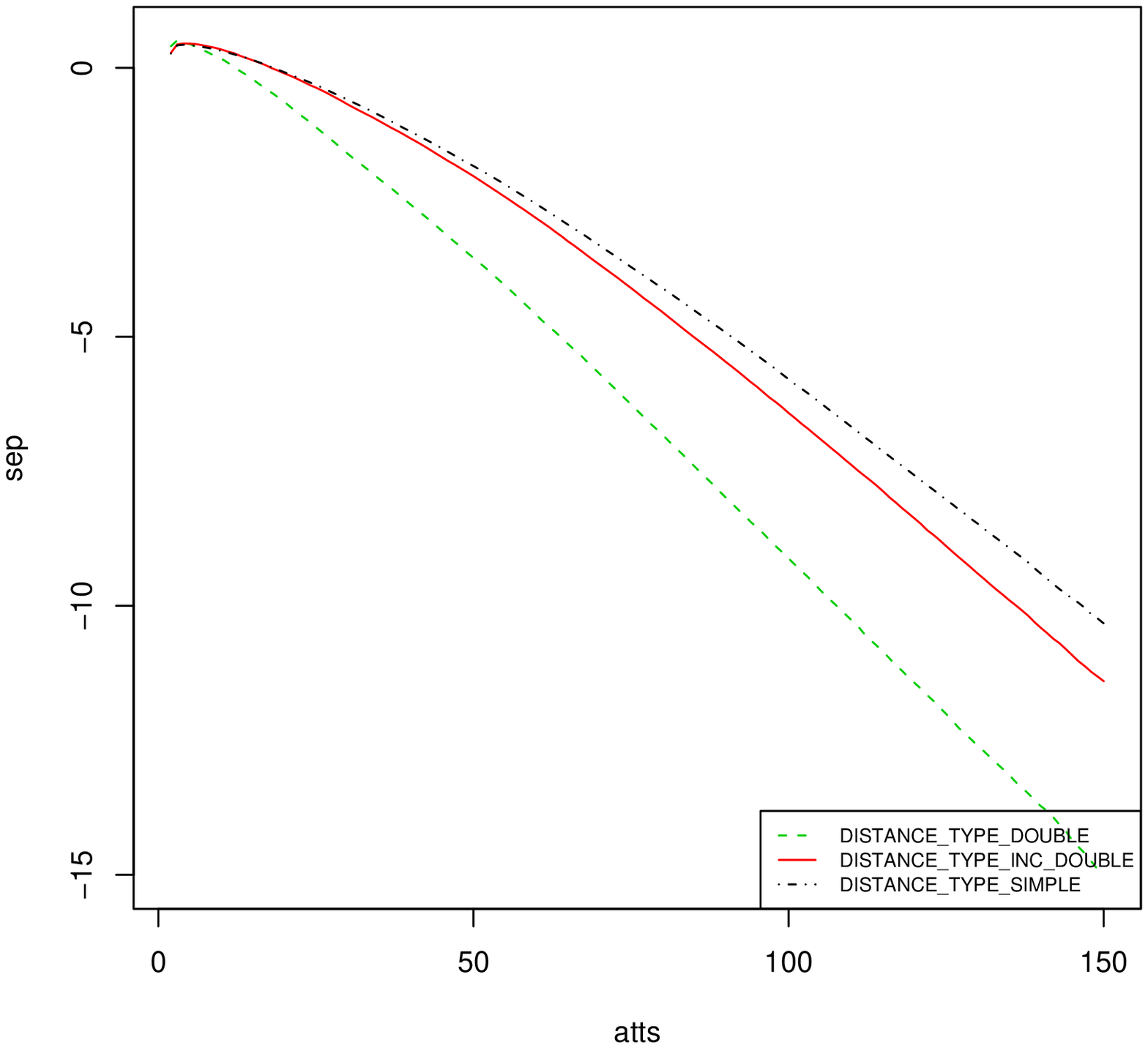}
	}}
	\subfigure[RandomRBFRandRed]{
	\scalebox{0.3}{
	\includegraphics{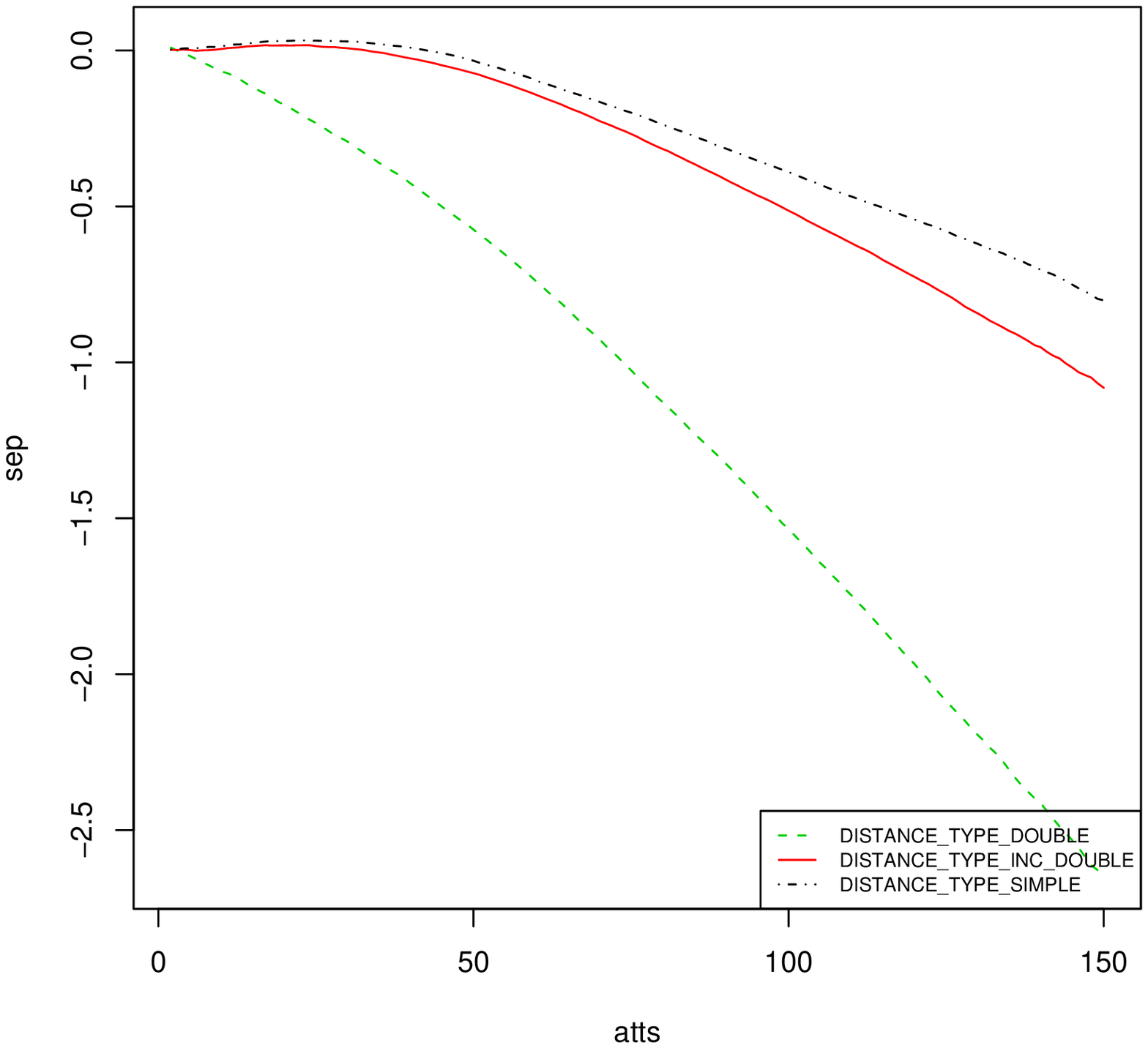}
	}}	
    }
    \mbox{
	\subfigure[NonMonotonic]{
	\scalebox{0.3}{
	\includegraphics{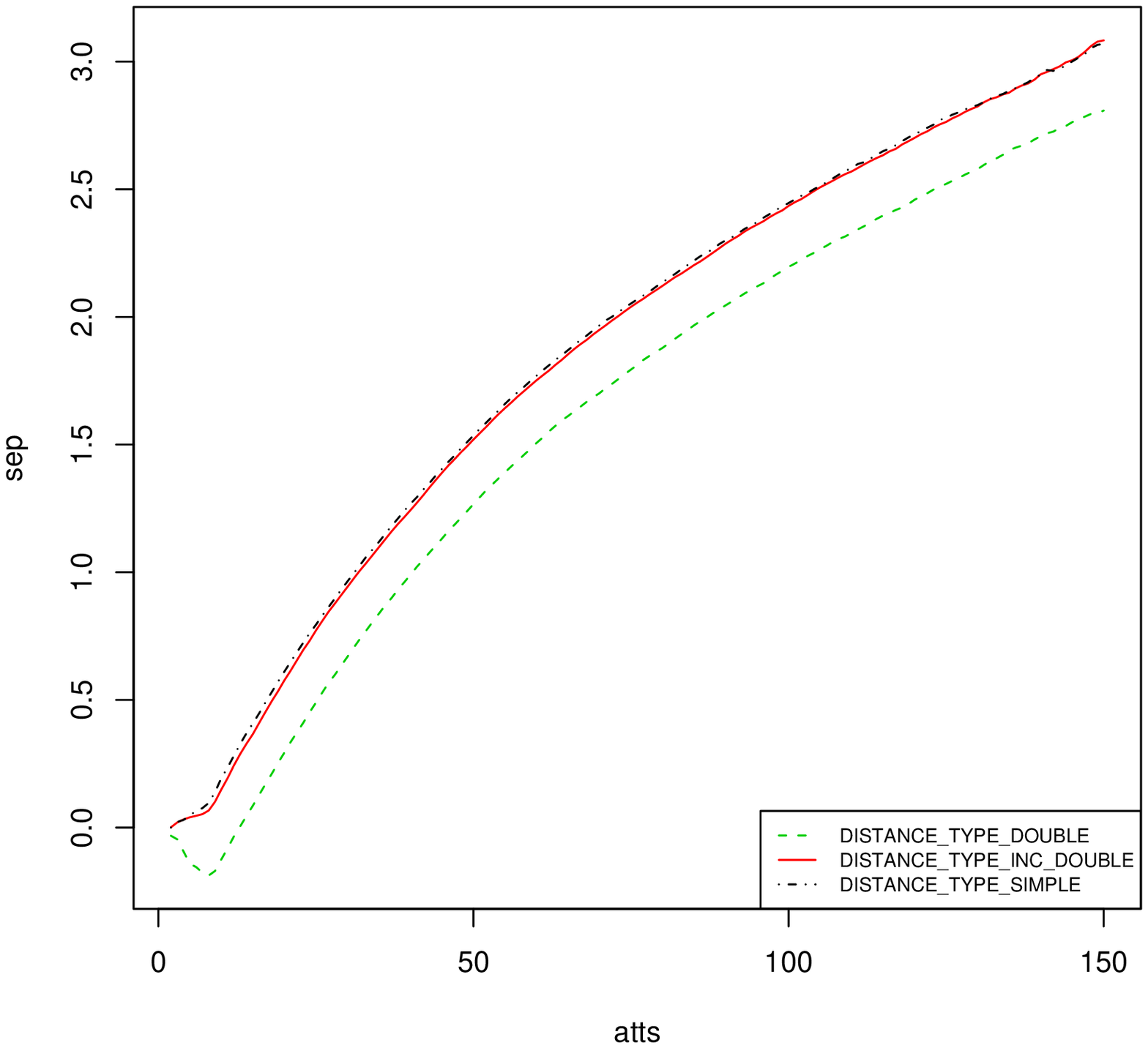}
	}}	
	\subfigure[MajorityN]{
	\scalebox{0.3}{
	\includegraphics{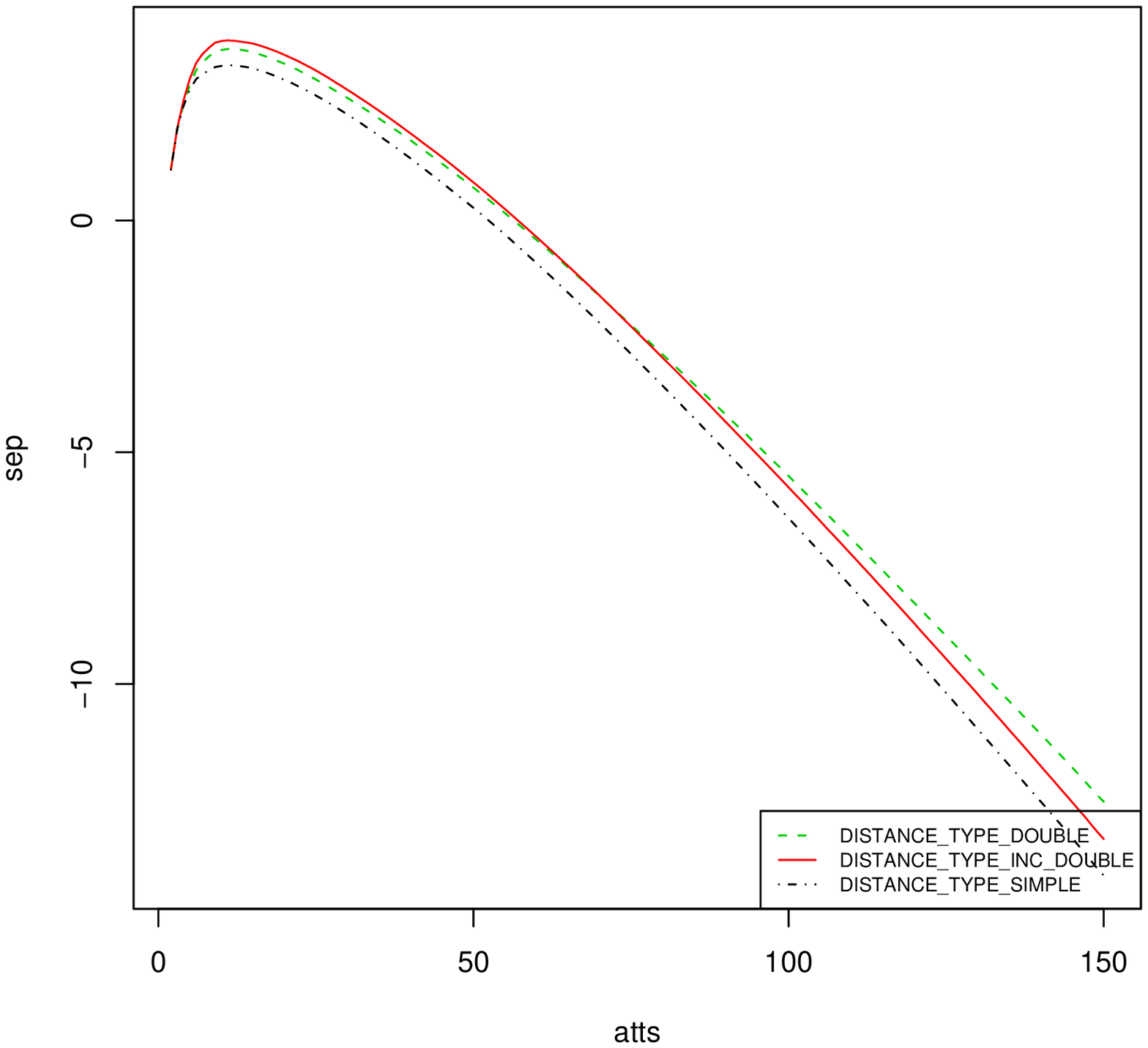}
	}}
    }
    \mbox{
	\subfigure[ModuloP]{
	\scalebox{0.3}{
	\includegraphics{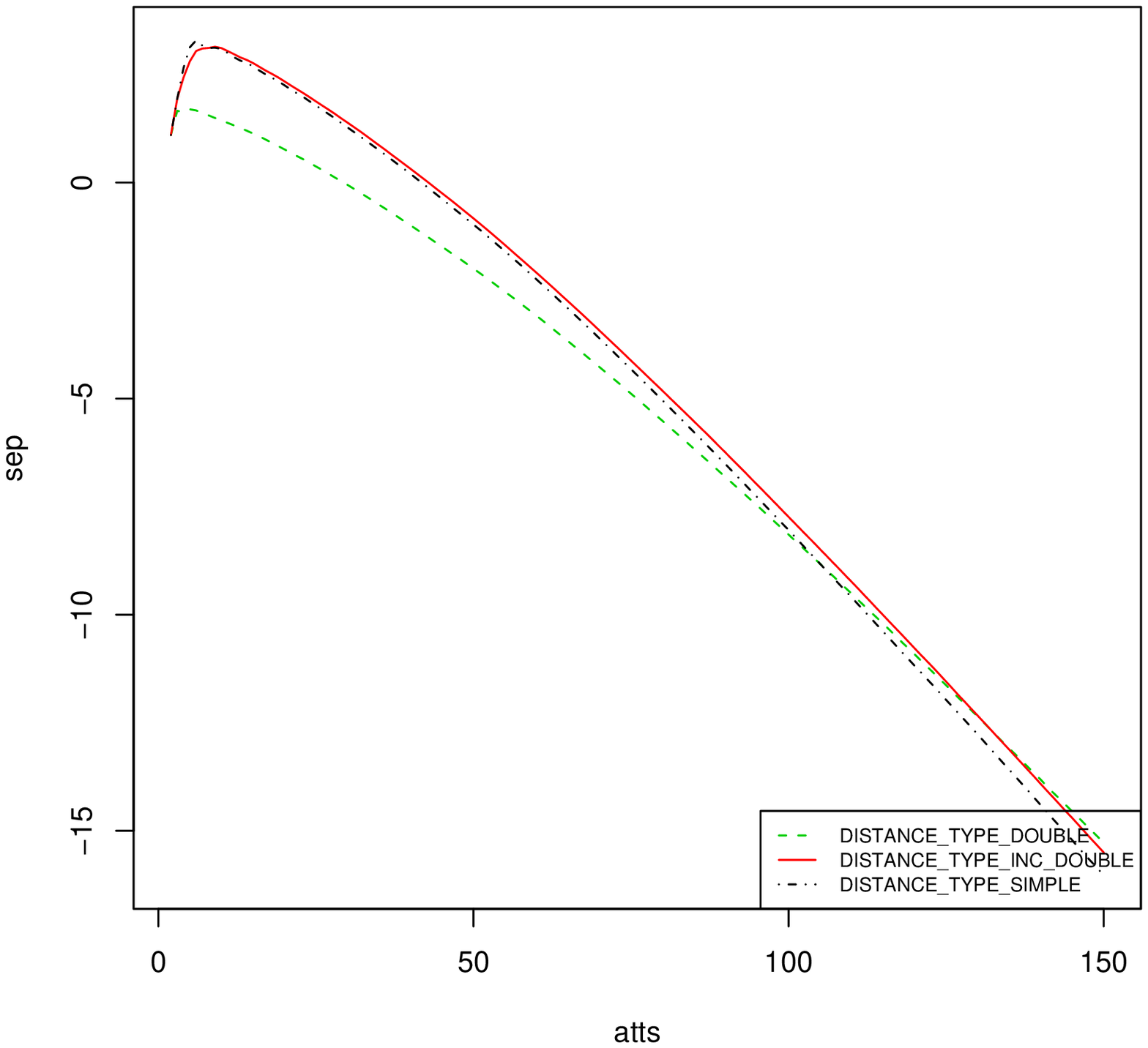}
	}}	
	\subfigure[RDG1NamedCategorical]{
	\scalebox{0.3}{
	\includegraphics{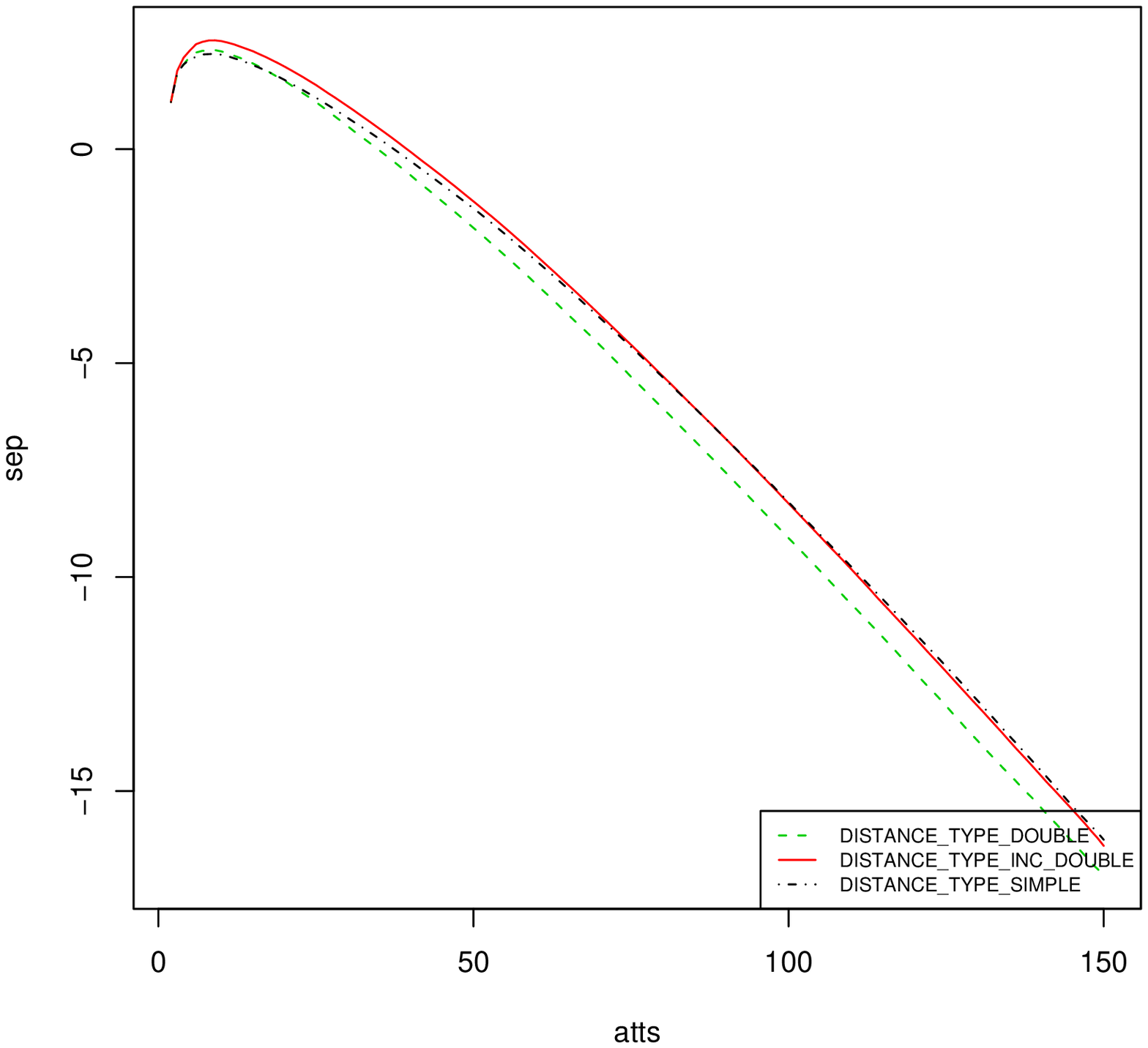}
	}}
    }
  \end{center}
  \caption {Accumulated separability versus total number of attributes for the three algorithms.}
  \label{fig:results_acu}
\end{figure}

Now, in order to accentuate the global differences between the three algorithms six more plots are presented with the accumulated results for the y-axis. Fig. \ref{fig:results_acu} shows these results. Now the x-axis keeps the same definition as before while the y-axis is the accumulated value of the separability, so now the formula for the y-axis value at point $x_n$ is the one in Eq. \ref{eq:results_acu} knowing that $s_i$ is the separability defined in Eq. \ref{eq:results} at point $x_i$.
\begin{equation}
\begin{gathered}
\text{y-axis: } \sum\limits_{i=0}^n{s_i} \hfill
\end{gathered}
\label{eq:results_acu}
\end{equation}
For this new axis definition, the slope of the function indicates positive or negative separability. If function descends at some point then separability was negative, on the other hand if function is ascending at this point then separability was positive. The steepness of the slope indicates the magnitude of the separability (either if it was positive or negative). And finally the separation between the curves for each algorithm tells about the accumulated difference of separabilities. If at the end one algorithm is above another it shows that the accumulated (and so the mean) separability is greater for this particular algorithm so one can conclude that in average this algorithm outperforms the other.

\section{Conclusions}\label{seq:conclusions}
By looking at the results above, it can be seen that none of the three algorithms is clearly better than another for the chosen set of problems. Looking at the first set of plots having separability is in the x-axis, we can see that the curves for three algorithms are almost the same, only when there are few attributes dReliefF seems to have different behavior. 

An anomaly is the problem of the random RBFs, there dReliefF is clearly worse. In fatct, except for the majority problem dReliefF is always the worse algorithm and even there it is non-significantly better. A difference between dReliefF and the other two algorithms is that it uses the calculated weights as distance ponderations starting at the first iteration of the algorithm. That certainly may cause ReliefF to get stuck into a local minimum found in those first iterations because the distance function that is using does not take into account some of the relevant variables. In a section above where pdReliefF is introduced, we stated the hypothesis that using the weights estimates since the first iteration may cause decrease performance due to the fact that these estimations may be too biased to the first instances and, so, may be far from the optimal weights. Now the results help support this hypothesis. That could also explain why dReliefF's behavior is different from the others when few attributes are evaluated as opposed as when more attributes are present. When there are few attributes to calculate distance with, making a mistake on choosing their ponderations makes big changes in the results, so problems with few attributes are more sensible to wrong distance calculations and cause dReliefF to either have much higher or lower performance depending on how close are the early weights to the real optimal weights. If the first instances seen by the algorithm are not representative of the whole set, for example because they share some common characteristic that is rare among other instances, then the weights used will be biased; on the other hand if these first instances give more accurate weight approximates, then is possible that dReliefF's worked better than the rest. 

There is also another characteristic of the results to be pointed out. In the second set of plots where differences among the algorithms stand out clearer, one can see differences between the behavior of the normal version of the algorithm as opposed to the modified ones. In these plots, two parallel curves for the separability of two algorithms, indicate that their performance evolves in the same way, meanwhile divergent curves indicate that the performance of one of them increases (decreases) more than the other. Having this in mind the results show that for the two first problems with numeric attributes the performance of dReliefF decreases very quick, normal ReliefF is the best of the three and pdReliefF is close to it though its performance also decreases faster than normal ReliefF's. Results for NonMonotonic are not clear as separability for that particular problem keeps very high for any number of attributes and the three algorithms perform almost identical. Some modifications could be applied to the generation of the problem to make it more difficult for ReliefF to discriminate attributes' relevance (e.g. adding more noise to the relevant ones) and compare the performance degradation for the three algorithms. The odd thing is that on the contrary of what happens with numeric problems, when we move onto the categoric ones we can see that now the algorithm which suffers the least performance decrease is dReliefF followed by pdReliefF.

So the final conclusion looking at these experimental results must be that although the performance of the three algorithms is frequently almost the same, the new algorithm pdReliefF introduced seems to be always in the middle of the other two quite stick to the better of the two while the other two are better or worse depending on the problem type, maybe depending on whether attributes are numeric or categoric. And also that dReliefF is very sensible to early errors on weight approximation of ReliefF so it must be used carefully.

As future work, more problems could be tested and specific experiments should be conducted to get deeper in the hypothesis that the different versions of ReliefF perform different on problems with numeric or categoric attributes. Also some tests on real data should be done using different classifiers to contrast them to the results on artificially generated ones.  

\bibliography{relief}	
	
\end{document}